\documentclass{article}

    \usepackage[preprint]{neurips_2019}

\usepackage{natbib}
\usepackage[utf8]{inputenc} % allow utf-8 input
\usepackage[T1]{fontenc}    % use 8-bit T1 fonts
\usepackage{hyperref}       % hyperlinks
\usepackage{url}            % simple URL typesetting
\usepackage{booktabs}       % professional-quality tables
\usepackage{amsfonts}       % blackboard math symbols
\usepackage{nicefrac}       % compact symbols for 1/2, etc.
\usepackage{microtype}      % microtypography
\usepackage[pdftex]{graphicx}
\title{Learning to estimate label uncertainty for automatic radiology report parsing}

\author{%
  Tobi Olatunji,  Li Yao \\
  Enlitic Inc\\
  10 Jackson St\\
  San Francisco, CA 94124 \\
  \texttt{tobi@enlitic.com, li@enlitic.com} \\
}

\begin{document}

\maketitle

\begin{abstract}
  Bootstrapping labels from radiology reports has become the scalable alternative to provide inexpensive ground truth for medical imaging. Because of the domain specific nature, state-of-the-art report labeling tools are predominantly rule-based. These tools, however, typically yield a binary 0 or 1 prediction that indicates the presence or absence of abnormalities. These hard targets are then used as ground truth to train image models in the downstream, forcing models to express high degree of certainty even on cases where specificity is low. This could negatively impact the statistical efficiency of image models. We address such an issue by training a Bidirectional Long-Short Term Memory Network to augment heuristic-based discrete labels of X-ray reports from all body regions and achieve performance comparable or better than domain-specific NLP, but with additional uncertainty estimates which enable finer downstream image model training.
\end{abstract}

\section{Introduction and Related Work}
X-rays are among the most prevalent imaging modalities in medical diagnosis. Consequently, most deep learning medical imaging applications detect anomalies on X-ray images \citep{wang2017clinical, yao2017learning, yao2018weakly, yao2019strong}. However, expert radiologists agree that X-ray is one of the least specific imaging modalities for clinical diagnosis when compared with other imaging modalities such as MRI and CT \citep{Smith2016ACO}. As a result, X-ray radiology reports inherently express a high degree of uncertainty.

Nevertheless, most traditional natural language processing (NLP) or rule-based systems extract labels from reports yielding dichotomous output for the presence (1) or absence (0) of abnormalities without mechanisms to express the associated degree of uncertainty \citep{olatunji2019caveats} \citep{hassanpour2016information} \cite{attaluri2018efficient}. At best, tools like NegBio \citep{peng2018negbio}, CheXpert labeller \citep{rajpurkar2017chexnet} and cTakes \citep{savova2010mayo} output a third class representing "uncertainty". When these hard targets are used downstream to train image models, it forces models to make a definitive prediction on all cases regardless of the confidence in the original radiology reports. This may lead to sub-optimal performance \citep{hinton2015distilling}. 

Inspired by \cite{hinton2015distilling}, we address the aforementioned issue by training sentence-based and report-based Long-Short Term Memory Networks (LSTMs) to augment discrete labels generated from the rule-based system with a continuous score which in turn may be interpreted as model's uncertainty or confidence. And we do so without sacrificing sensitivity and specificity.

In particular, a rule-based in-house NLP tool is used to first classify a report into either normal or abnormal. Given such discrete binary labels, LSTMs are then trained to reproduce them. As the by-product of training, the continuous predictions from LSTMs may be used to capture the confidence and uncertainty of a binary prediction.

\section{Experiments}

\subsection{Datasets} 
For training, we use a private dataset that covers 6 body regions (abdomen, chest, spine, upper extremity, lower extremity and head/neck), a total of about 900,000 reports. For testing, we use two datasets, one public and one private. The public dataset from OpenI consists of 7,468 chest X-ray reports along with their ground truth labels \citep{demner2012design} while the private dataset had 2,185 reports hand-labelled by 3 expert radiologists. 

%The training set had 71\% abnormal reports, while public OpenI test set has 67\% abnormal ones.

\subsection{NLP labeling} 
We extract labels from the reports using domain-specific rule-based NLP tools. We developed the NLP tool in 3 steps. (1) Extraction: we extract findings in the report using NIH's METAMAP \citep{aronson2010overview} adding further heuristics to improve sensitivity and specificity. (2) Negation detection: We craft negation rules based on the output of Stanford's CoreNLP dependency parser \citep{manning-EtAl:2014:P14-5}. (3) Classification: we craft rules to filter findings based on negation detection results and return a global label (normal/abnormal) for each report and for each report sentence.

\subsection{LSTM training} 
With each sentence and its binary label, we train a Bidirectional LSTM \citep{hochreiter1997long} from scratch in Keras using Tensorflow backend. The embedding layer is a matrix of 100 (embedding dimension) by 22,000 (vocabulary size), followed by a 1D spatial dropout layer of 0.2. This was followed by a BiLSTM layer with 256 hidden units and recurrent dropout of 0.4 for regularization. A dense layer with sigmoid activation then outputs model predictions. Training minimizes the binary cross entropy loss using adaptive moment \citep{kingma2014adam} as the optimizer with an initial learning rate of 0.001, beta1 as 0.9, beta2 as 0.999, and epsilon as 1e-07. We train on 8 Tesla V100 GPUS with minibatch size of 32 samples over 20 epochs using a patience of 5 for early stopping. At test time, we ensemble the predictions for each sentence in a report by taking the maximum (maxpooling) and compare this against ground truth report labels. %\textcolor{red}{Describe the training in more details}
%train on 90\% of the data and validate on 10\%. We compare baseline model with versions with pre-trained word2vec embeddings and Phraser model from gensim for multi-word embedddings.

\section{Results}

\begin{figure}[h]
\begin{center}
\includegraphics[width=1.0\linewidth]{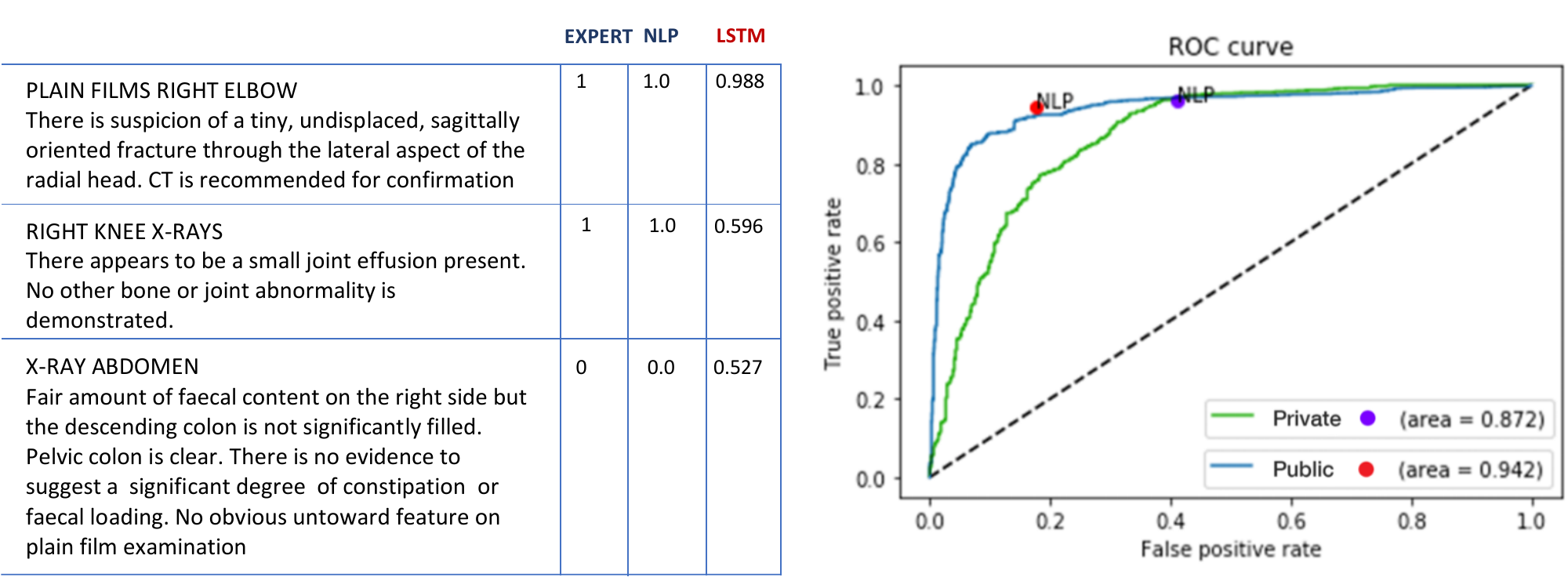}
\end{center}
\caption{\textbf{Left:} Examples of uncertainty in reports. \textbf{Right:} BiLSTM Performance compared with Rule-based NLP single operating point on Public (OpenI) and Private Datasets}
\end{figure}

In the first example above (left top), dichotomous labels sometimes mislead model prediction. However, the second report (left middle) and third report (left bottom) show reports where the report uncertainty is retained despite binary targets. The model instead produces uncertainty estimates, soft targets, instead of hard binary targets that serve as a confidence score otherwise unavailable to downstream image models. 

\section{Discussion: major implications of using soft labels} 

\paragraph{Soft target as a regularizer} Soft targets effectively scale the learning gradients, resulting in smoother updates to model weights, and increasing the models' robustness to labeling noise \citep{hinton2015distilling}.

\paragraph{Application-specific thresholding} With a continuous prediction, sensitivity and specificity can be adjusted based on specific use cases. For instance, a good triage model would select a threshold that focuses on achieving high sensitivity instead of specificity.

\paragraph{Study prioritization} Unlike binary predictions, uncertainty estimates naturally enable case prioritization in a clinical environment. Abnormal cases with high confidence may be reviewed in a more timely fashion. On the other hand, cases with high uncertainty may be diverted to more experienced clinicians for better diagnosis.
\paragraph{Labeling Efficiency} inherent complexity of rule-based systems make them less efficient at test time when compared with model predictions. For context, our rule-based system takes an average of 254 seconds to process 1000 reports using 80 2.5GHz CPU cores. By contrast, it took about 239 secs to label 7486 openi reports using 8 Tesla V100 GPUS. On our dataset of 900,000 reports, that could be a 10x speedup.

\bibliography{neurips_2019.bib}
\bibliographystyle{plainnat.bst}
\medskip

\end{document}